%% file: acl.tex
\pdfoutput=1

\documentclass[11pt]{article}

\usepackage{acl}

\usepackage{times}
\usepackage{latexsym}

\usepackage[T1]{fontenc}

\usepackage[utf8]{inputenc}

\usepackage{booktabs}
\usepackage{multirow}
\usepackage{amsmath}
\usepackage{amssymb}
\usepackage{amsfonts}
\usepackage{graphicx}
\usepackage{arydshln}
\usepackage{enumitem}
\usepackage{gensymb}

\usepackage[table]{colortbl}%
\usepackage{xcolor}

\usepackage{soul}
\usepackage{subcaption}

\usepackage{microtype}

\usepackage{cleveref}
\crefname{section}{\S}{\S\S}
\crefname{table}{Tab.}{}
\crefname{figure}{Fig.}{}
\crefname{algorithm}{Alg.}{}
\crefname{equation}{Eq.}{}
\crefname{appendix}{App.}{}
\crefformat{section}{\S#2#1#3}  %

\definecolor{muchworse}{RGB}{255, 100, 100} %

\definecolor{muchbetter}{RGB}{100, 255, 100} %

\definecolor{worse}{RGB}{255, 200, 200} %

\definecolor{better}{RGB}{200, 255, 200} %

\newcommand{\src}{\theta_{src}}
\newcommand{\tgt}{\theta_{tgt}}
\newcommand{\bilingual}{\theta_{\{src,tgt\}}}
\newcommand\blfootnote[1]{%
  \begingroup
  \renewcommand\thefootnote{}\footnote{#1}%
  \addtocounter{footnote}{-1}%
  \endgroup
}

\title{Zero-shot Cross-lingual Transfer is Under-specified Optimization}

\author{Shijie Wu, Benjamin Van Durme, Mark Dredze \\
Department of Computer Science \\
Johns Hopkins University \\
{\tt shijie.wu@jhu.edu, vandurme@jhu.edu, mdredze@cs.jhu.edu}
}

\begin{document}
\maketitle
\begin{abstract}
Pretrained multilingual encoders enable zero-shot cross-lingual transfer, but often produce unreliable models that exhibit high performance variance on the target language. We postulate that this high variance results from \textit{zero-shot cross-lingual transfer solving an under-specified optimization problem}. We show that any linear-interpolated model between the source language monolingual model and source + target bilingual model has equally low source language generalization error, yet the target language generalization error reduces smoothly and linearly as we move from the monolingual to bilingual model, suggesting that the model struggles to identify good solutions for both source and target languages using the source language alone. Additionally, we show that zero-shot solution lies in non-flat region of target language error generalization surface, causing the high variance.
\blfootnote{Code is available at \url{https://github.com/shijie-wu/crosslingual-nlp}.}
\end{abstract}

\section{Introduction}
Pretrained multilingual encoders like Multilingual BERT \citep[mBERT;][]{devlin-etal-2019-bert} and XLM-RoBERTa \citep[XLM-R;][]{conneau-etal-2020-unsupervised} facilitate zero-shot cross-lingual transfer \citep{wu-dredze-2019-beto,hu2020xtreme} --- training the model on one language then using it on another language without additional task-specific training data.
While the generalization performance on the source language has low variance, on the target language the variance is much higher with zero-shot cross-lingual transfer \citep{keung-etal-2020-dont,wu-dredze-2020-explicit}, making it difficult to compare different models in the literature. Similarly, pretrained monolingual encoders also have unstable performance during fine-tuning \citep{devlin-etal-2019-bert,phang2018sentence}.

Why are these models so sensitive to the random seed? Many theories have been offered: catastrophic forgetting of the pretrained task \citep{phang2018sentence,Lee2020Mixout,keung-etal-2020-dont}, 
small data size \citep{devlin-etal-2019-bert}, 
impact of random seed on task-specific layer initialization and data ordering \citep{dodge2020fine}, the Adam optimizer without bias correction \citep{mosbach2021on,zhang2021revisiting}, and a different generalization error with similar training loss \citep{mosbach2021on}. However, none of these factors fully explain the high generalization error variance of zero-shot cross-lingual transfer on target language but low variance on source language.

We offer a new explanation for high variance in target language performance: \textit{the zero-shot cross-lingual transfer optimization problem is under-specified}. 
Based on the well-established linear interpolation of 1-dimensional plot and contour plot \citep{goodfellow2014qualitatively,li2018visualizing}, we empirically show that any linear-interpolated model between the monolingual source model and bilingual source and target model has equally low source language generation error. Yet the target language generation error surprisingly reduces smoothly and linearly as we move from a monolingual model to a bilingual model. To the best of our knowledge, no other paper documents this finding.

This result provides a new answer to our mystery: only a small subset of the solution space for the source language solves the target language on par with models with actual target language supervision; the optimization could not find such a solution with existing condition (without target language supervision), hence an under-specified optimization problem. If target language supervision were available, as it was in the counterfactual bilingual model, the optimization would find the smaller subset. By comparing both mBERT and XLM-R, we find that the generalization error surface of XLM-R is flatter than mBERT, contributing to its better performance compared to mBERT. Thus, zero-shot cross-lingual transfer has high variance, as the solution found by zero-shot cross-lingual transfer lies in the non-flat region of the target language generalization error surface. Small turbulence on the parameter space would lead to big generalization error difference, hence the high variance.

\section{Existing Hypotheses (Related Work)}
\label{sec:prior-work}
Prior studies have observed fine-tuning variance with pretrained encoder, and have offered various hypotheses to explain this behavior.
Catastrophic forgetting -- when neural networks trained on one task forget that task after training on a second task 
 \cite{McCloskey1989CatastrophicII,kirkpatrick2017overcoming}
---has been credited as the source of high variance in both monolingual fine-tuning \cite{phang2018sentence,Lee2020Mixout} and zero-shot cross-lingual transfer \cite{keung-etal-2020-dont}.  \citet{mosbach2021on} wonders why preserving cloze capability is important. However, in zero-shot cross-lingual transfer, deliberately preserving the multilingual cloze capability with regularization improves performance but does not eliminate the zero-shot transfer gap \cite{aghajanyan2021better,liu-etal-2021-preserving}.

Small training data size often seems to have higher variance in performance \cite{devlin-etal-2019-bert}, but \citet{mosbach2021on} found that when controlling the number of gradient updates, smaller data size has the similar variance as larger data size. 

In the pretraining-then-fine-tune paradigm, random seeds impact the initialization of task-specific layers and data ordering during fine-tuning. \citet{dodge2020fine} shows development set performance has high variance with respect to seeds. Additionally, Adam optimizer without bias correction---an Adam \citep{kingma2014adam} variant (inadvertently) introduced by the implementation of \citet{devlin-etal-2019-bert}---has been identified as the source of high variance during monolingual fine-tuning \citep{mosbach2021on,zhang2021revisiting}. However, in zero-shot cross-lingual transfer, while different random seeds lead to high variance in target languages, the source language has much smaller variance in comparison even with standard Adam \citep{wu-dredze-2020-explicit}.

Beyond optimizers, \citet{mosbach2021on} attributes high variance to generalization issues: despite having similar training loss, different models exhibit vastly different development set performance. However, in zero-shot cross-lingual transfer, the development or test performance variance is much smaller on the source language compared to target language.

\section{Under-specified Optimization}
\label{sec:analysis}

\begin{figure}[t]
\centering
\includegraphics[width=0.7\columnwidth]{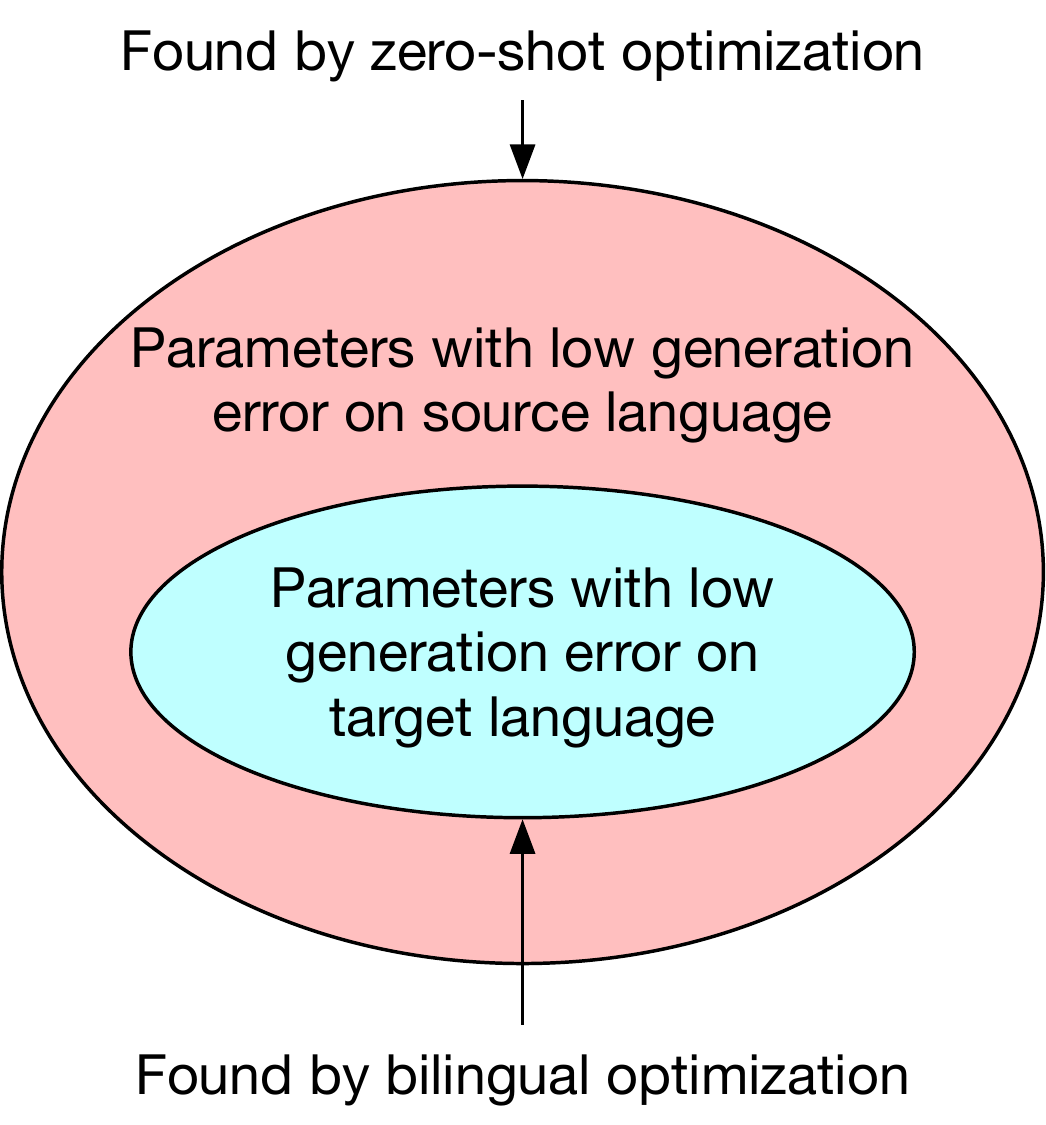}
\caption{zero-shot cross-lingual transfer is an under-specified optimization problem. With the existing condition, the optimization could not find the solution that we really want.}
\label{fig:venn}
\end{figure}

\begin{figure*}[t]
\centering
\includegraphics[width=2\columnwidth]{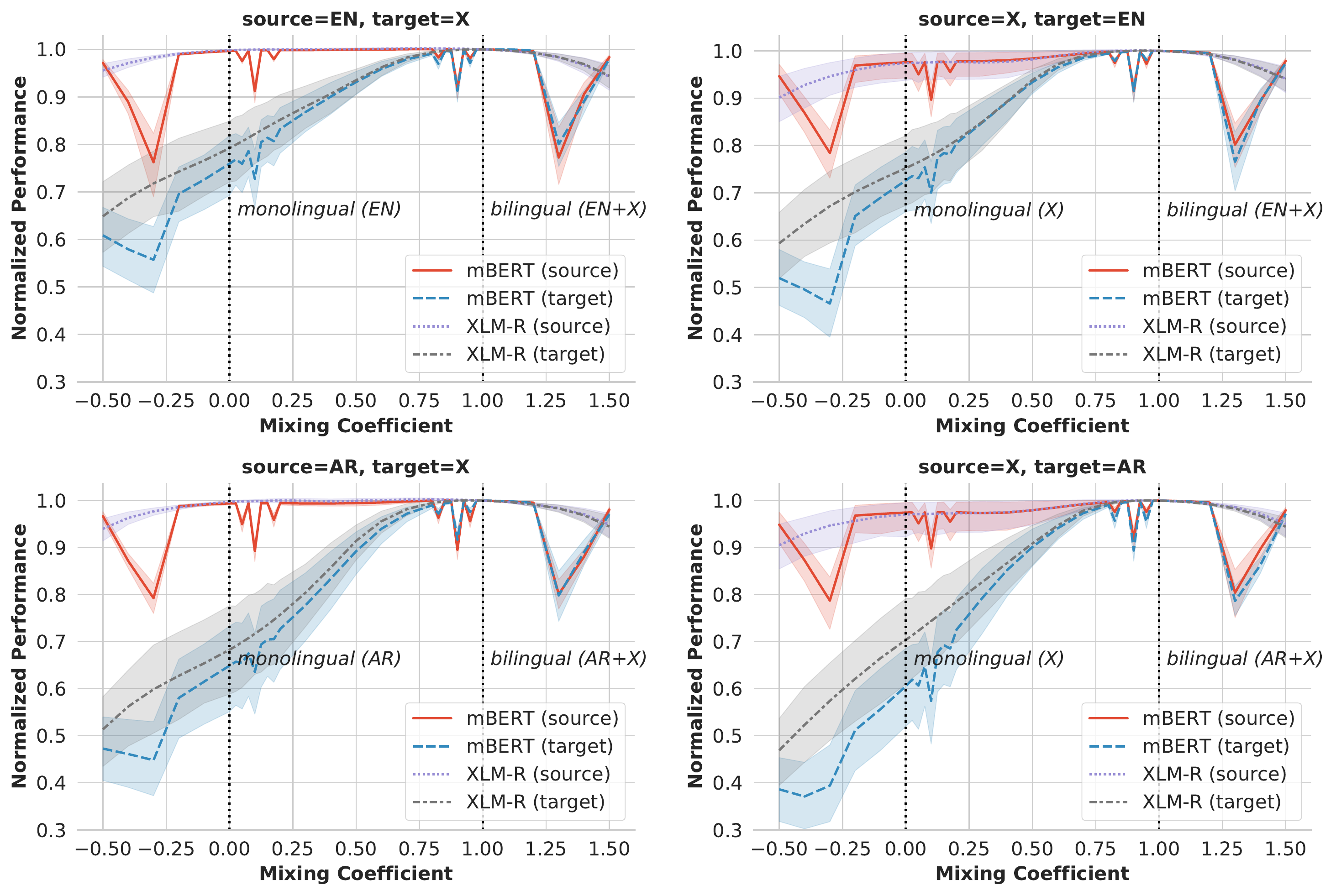}
\caption{Normalized performance of a linear interpolated model between a monolingual and bilingual model. A single plot line shows the performance normalized by the matching bilingual model and aggregated over eight language pairs and four tasks, with the shaded region representing 95\% confidence interval. 
The x-axis is the linear mixing coefficient $\alpha$ in \cref{eq:src-bi} and \cref{eq:tgt-bi}, with $\alpha=0$ and $\alpha=1$ representing source language monolingual model and source + target bilingual model, respectively.
Each subfigure title indicates the source and target languages.
Across all experiments, the source language dev performance stays consistently high (red and purple lines) during interpolation while the target language dev performance starts low and increases smoothly and linearly as it moves towards the bilingual model (gray and blue lines).
\cref{sec:mean-breakdown} break down this figure by tasks.}
\label{fig:interpolation-mean}
\end{figure*}

\begin{figure*}[t]
\centering
\subfloat[][EN-RU NER w/ mBERT]{
\includegraphics[width=0.5\columnwidth]{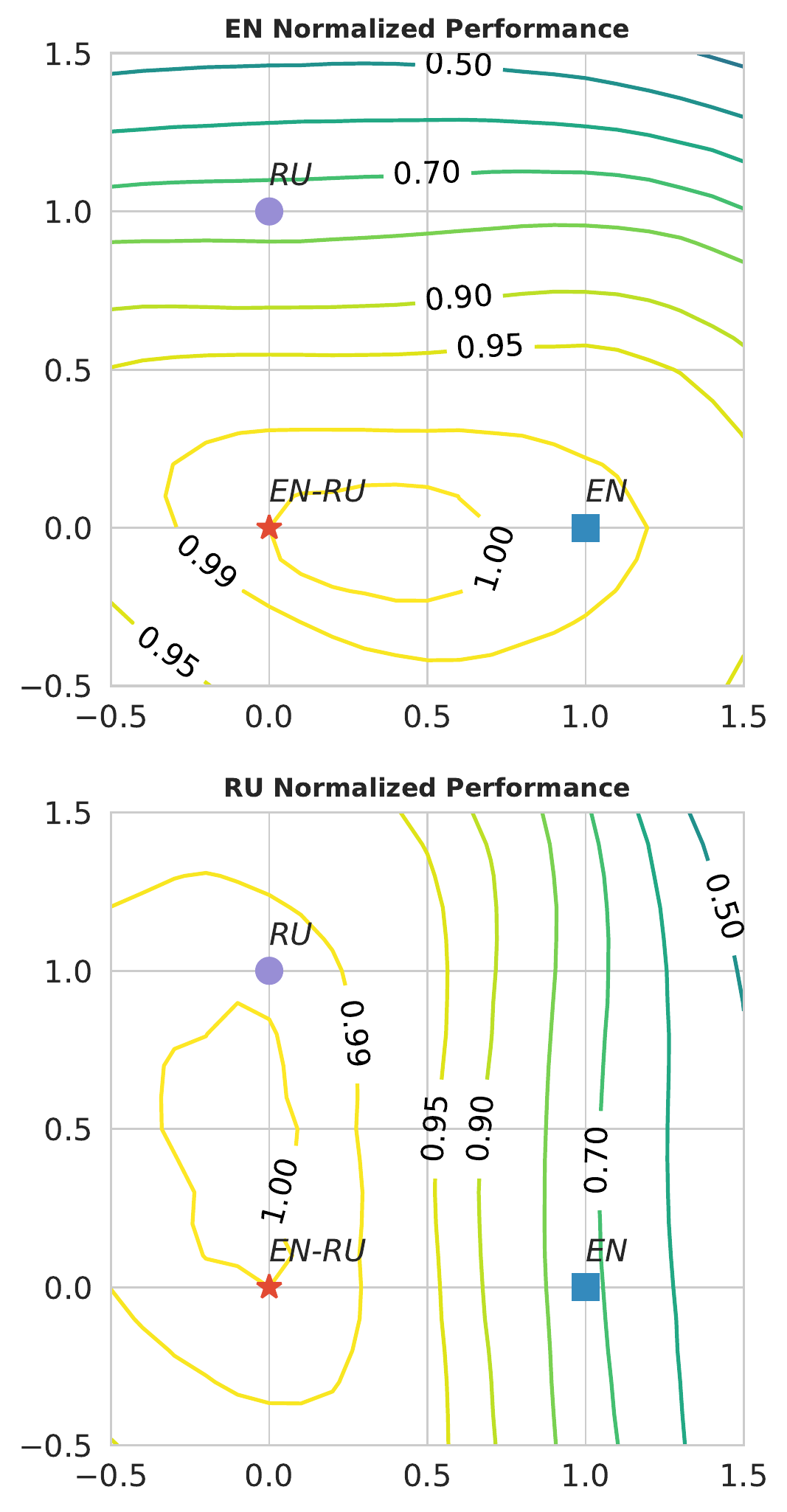}
}
\subfloat[][EN-RU NER w/ XLM-R]{
\includegraphics[width=0.5\columnwidth]{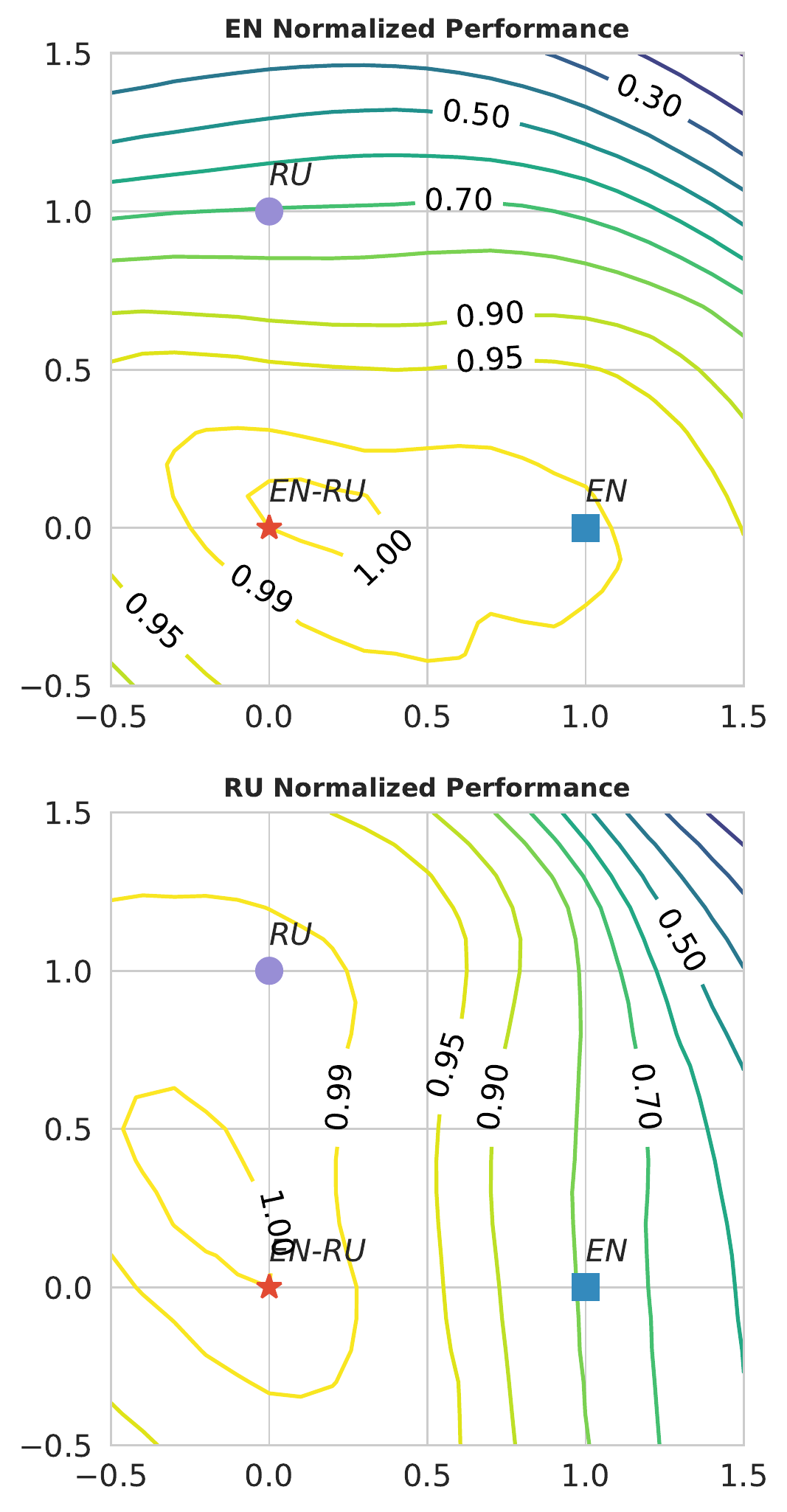}
}
\subfloat[][AR-ZH XNLI w/ mBERT]{
\includegraphics[width=0.5\columnwidth]{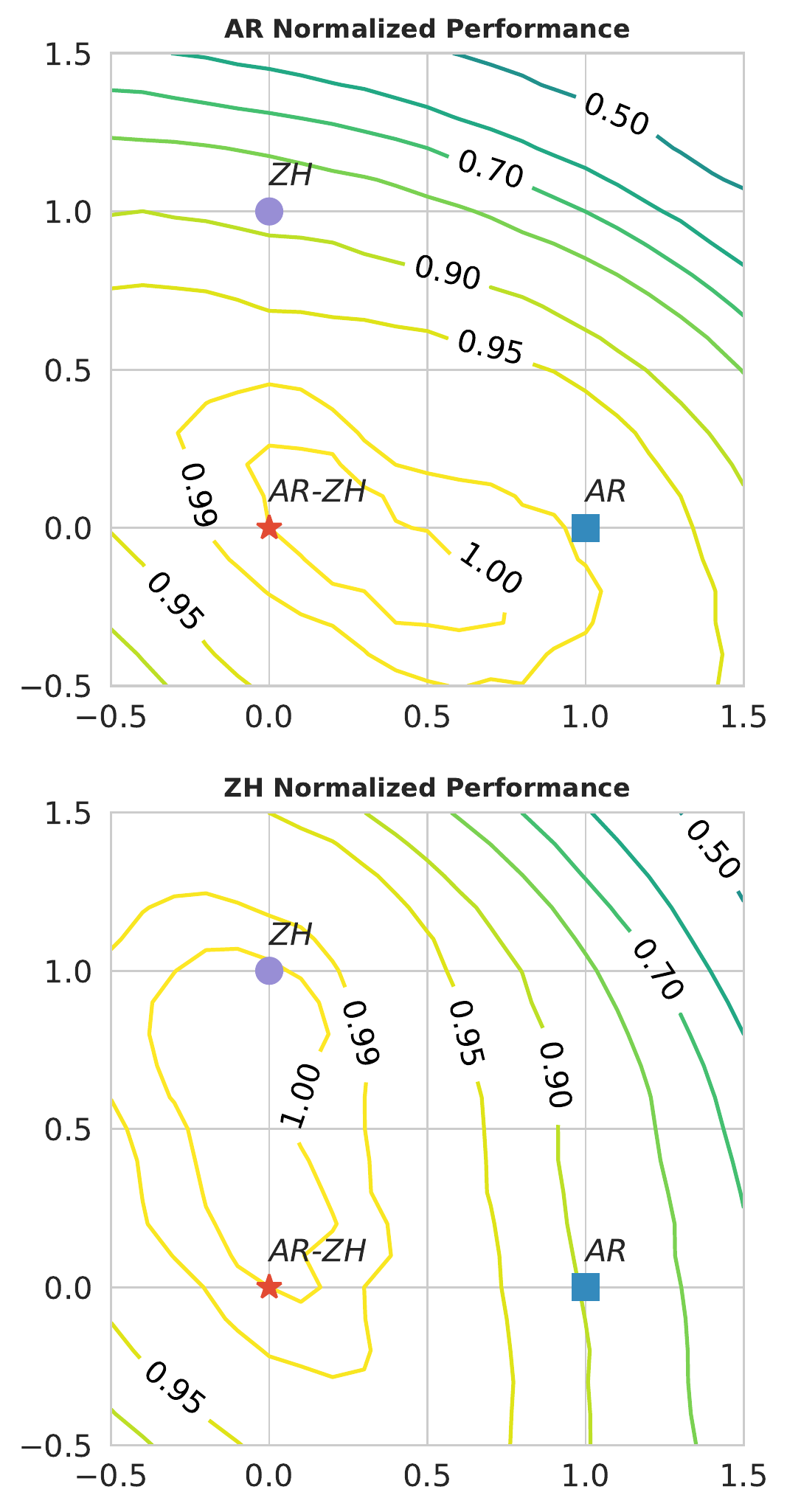}
}
\subfloat[][AR-ZH XNLI w/ XLM-R]{
\includegraphics[width=0.5\columnwidth]{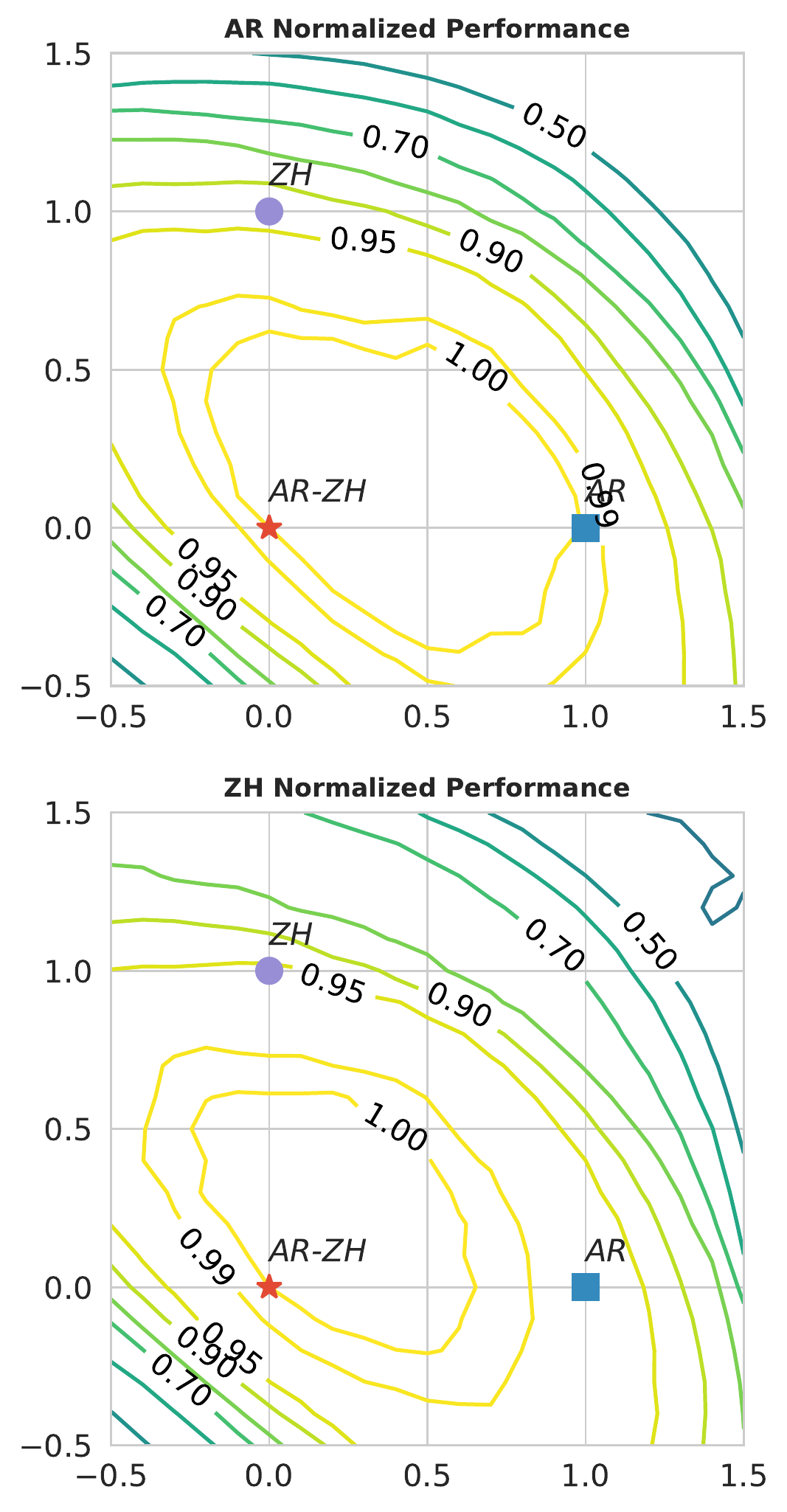}
}

\caption{Normalized performance of 2D linear interpolation between bilingual model and monolingual models.
The x-axis and the y-axis are the $\alpha_1$ and $\alpha_2$ in \cref{eq:tri}, respectively.
By comparing mBERT and XLM-R, we observe that XLM-R has a flatter target language generalization error surface compared to mBERT.
Different language pairs and tasks combinations show similar trends and additional figures can be found in \cref{sec:interpolation-2d-others}
}
\label{fig:interpolation-2d}
\end{figure*}
Existing hypotheses do not explain the high variance of zero-shot cross-lingual transfer:
much higher variance on generalization error of the target language compared to the source language. 
We propose a new explanation: \textit{zero-shot cross-lingual transfer is an under-specified optimization problem}.\footnote{This explanation provides deeper insight on the common belief that no target data causes high variance. We provide evidence on how these two factors interact.} 
As in \cref{fig:venn}, optimizing a multilingual model for a specific task using only source language annotation allows choices of many good solutions in terms of generalization error. However, unbeknownst to the optimizer, these solutions have wildly different generalization errors on the target language. In fact, a small subset has similar low generalization error as models trained on target language. Yet without the guidance of target data, the zero-shot cross-lingual optimization could not find this smaller subset. As we will show in \cref{sec:result}, the solution found by zero-shot transfer lies in a non-flat region of target language generalization error, and small turbulence in the parameter space causes a big difference in generalization error, causing its high variance.

\subsection{Linear Interpolation}\label{sec:interpolation}
We test this hypothesis via a linear interpolation between two models to explore the neural network parameter space.
Consider three sets of neural network parameters: $\src$, $\tgt$, $\bilingual$ for a model trained on task data for the source language only, target language only and both languages, respectively. This includes both task-specific layers and encoders.\footnote{We also experiment with interpolating the encoder parameters only and observe similar findings. On the other hand, interpolating the task-specific layer only has a negligible effect.}
Note all three models have the same initialization before fine-tuning, making the bilingual model a counterfactual setup if the corresponding target language supervision were available.
We obtain the 1-dimensional (1D) linear interpolation of a monolingual (source) task trained model and bilingual task trained model with
\begin{equation}\label{eq:src-bi}
\theta(\alpha) = \alpha\bilingual + (1-\alpha) \src
\end{equation}
or we could swap source and target by
\begin{equation}\label{eq:tgt-bi}
\theta(\alpha) = \alpha\bilingual + (1-\alpha) \tgt
\end{equation}
where $\alpha$ is a scalar mixing coefficient \cite{goodfellow2014qualitatively}. 
Additionally, we can compute a 2-dimensional linear interpolation as
\begin{equation}\label{eq:tri}
\theta(\alpha_1, \alpha_2) = \bilingual + \alpha_1 \delta_{src} + \alpha_2 \delta_{tgt} 
\end{equation}
where $\delta_{src} = \src-\bilingual$, $\delta_{tgt} = \tgt-\bilingual$, $\alpha_1$ and $\alpha_2$ are scalar mixing coefficients \cite{li2018visualizing}.\footnote{\citet{li2018visualizing} use two random directions and they normalize it to compensate scaling issue. In this setup, we find $\delta_{src}$ and $\delta_{tgt}$ have near identical norms, so we do not apply additional normalization. As these two directions are not random, we find that it spans around 55\degree. We plot the norm ratio and angle of these two vectors in \cref{sec:dist-vs-angle}.} Finally, we can evaluate any interpolated models on the development set of source and target languages, testing the generalization error on the same language and across languages. 

The performance of the interpolated model illuminates the behavior of the model's parameters.
Take \cref{eq:src-bi} as an example: if the linear interpolated model performs consistently high for our task on the source language, it suggests that both models lie within the same local minimum of source language generalization error surface. Additionally, if the linear interpolated model performs vastly differently on the target language, it would support our hypothesis. On the other hand, if the linear interpolated model performance drops on the source language, it suggests that both models lie in different local minimum of source language generalization error surface, suggesting zero-shot optimization searching the wrong region.

\section{Experiments}

We consider four tasks: natural language inference \citep[XNLI;][]{conneau-etal-2018-xnli}, named entity recognition \citep[NER;][]{pan-etal-2017-cross}, POS tagging and dependency parsing \citep{ud2.7}.
We evaluate XNLI and POS tagging with accuracy (ACC), NER with span-level F1, and parsing with labeled attachment score (LAS).
We consider two encoders: base mBERT and large XLM-R.
For the task-specific layer, we use a linear classifier for XNLI, NER, and POS tagging, and \citet{dozat2016deep} for dependency parsing. 

To avoid English-centric experiments, we consider two source languages: English and Arabic. We choose 8 topologically diverse target languages: Arabic\footnote{Arabic is only used when English is the source language.}, German, Spanish, French, Hindi, Russian, Vietnamese, and Chinese.
We train the source language only and target language only monolingual model as well as a source-target bilingual model. 

We compute the linear interpolated models as described in \cref{sec:interpolation} and test it on both the source and target language development set.
We loop over $\{-0.5,-0.4,\cdots,1.5\}$ for $\alpha$, $\alpha_1$ and $\alpha_2$.\footnote{We additionally select 0.025, 0.05, 0.075, 0.125, 0.15, 0.175, 0.825, 0.85, 0.875, 0.925, 0.95, and 0.975 for $\alpha$ due to preliminary experiment.}
We report the mean and variance of three runs by using different random seeds. We normalized both mean and variance of each interpolated model by the bilingual model performance, allowing us to aggregate across tasks and language pairs. Details of fine-tuning can be found in \cref{sec:fine-tuning}.

\section{Results} \label{sec:result}

In \cref{fig:interpolation-mean}, we observe that interpolations between the source monolingual and bilingual model have consistently similar source language performance. In contrast, surprisingly, the target language performance smoothly and linearly improves as the interpolated model moves from the zero-shot model to bilingual model.\footnote{We also show the variance of the interpolated models in \cref{sec:variance}. The source language has much lower variance compared to target language on the monolingual side of the interpolated models, echoing findings in \citet{wu-dredze-2020-explicit}.}
The only exception is mBERT, where the performance drops slightly around 0.1 and 0.9 locally. In contrast, XLM-R has a flatter slope and smoother interpolated models. 

\cref{fig:interpolation-2d} further demonstrates this finding with a 2D linear interpolation. The generalization error surface of the target language of XLM-R is much flatter compared to mBERT, perhaps the fundamental reason why XLM-R performs better than mBERT in zero-shot transfer, similar to findings in CV models \cite{li2018visualizing}. As we discuss in \cref{sec:analysis}, these two findings support our hypothesis that zero-shot cross-lingual transfer is an under-specified optimization problem. As \cref{fig:interpolation-2d} shows, the solution found by zero-shot transfer lies in a non-flat region of target language generalization error surface, causing the high variance of zero-shot transfer on the target language. In contrast, the same solution lies in a flat region of the source language generalization error surface, causing the low variance on the source language.

\section{Discussion}

We have presented evidence that zero-shot cross-lingual transfer is an under-specified optimization problem, and the cause of high variance on target language but not the source language tasks during cross-lingual transfer. This finding holds across 4 tasks, 2 source languages and 8 target languages.
Training bigger encoders addresses this issue indirectly by producing encoders with flatter cross-lingual generalization error surfaces. 
However, a more robust solution may be found by introducing constraints into the optimization problem. %
There are a few potential solutions.

Few-shot cross-lingual transfer is a potential way to further constrain the optimization problem. \citet{zhao-etal-2021-closer} finds that it is important to first train on source language then fine-tune with the few-shot target language examples. Through the lens of our analysis, this finding is intuitive since fine-tuning with a small amount of target data provides a guidance (gradient direction) to narrow down the solution space, leading to a potentially better solution for the target language. The initial fine-tuning with the source data is also important since it provides a good starting point. Additionally, \citet{zhao-etal-2021-closer} observes that the choice of shots matters. This is expected as it significantly impacts the quality of the gradient direction.

Similarly, silver target data is a potential way to further constrain the optimization problem. While \citet{yarmohammadi2021everything} finds that jointly training with gold source data and silver target data benefits cross-lingual transfer, a pipeline fine-tuning approach like few-shot cross-lingual transfer is also worth exploring.

Unsupervised model selection like \citet{chen2020model} and optimization regularization like \citet{aghajanyan2021better} have been proposed in the literature to improve zero-shot cross-lingual transfer. Through the lens of our analysis, both solutions attempt to constrain the optimization problem.

As none of the existing techniques fully constrain the optimization, future work should study the combination of existing techniques and develop new techniques on top of it instead of studying one technique at a time. We leave the exploration of this to future work.

\section*{Acknowledgments}
This research is supported in part by ODNI, IARPA, via the BETTER Program contract \#2019-19051600005. The views and conclusions contained herein are those of the authors and should not be interpreted as necessarily representing the official policies, either expressed or implied, of ODNI, IARPA, or the U.S. Government. The U.S. Government is authorized to reproduce and distribute reprints for governmental purposes notwithstanding any copyright annotation therein.

This research is supported by the following open-source softwares: NumPy \citep{2020NumPy-Array}, PyTorch \citep{paszke2017automatic}, PyTorch lightning \cite{falcon2019pytorch}, scikit-learn \citep{JMLR:v12:pedregosa11a}, Transformer \citep{Wolf2019HuggingFacesTS}.

\bibliography{anthology,custom}
\bibliographystyle{acl_natbib}

\clearpage

\appendix

\section{Fine-tuning Experiments Detail}\label{sec:fine-tuning}
We follow the implementation and hyperparameter of \citet{wu-dredze-2020-explicit}. We optimize with Adam \citep{kingma2014adam}. The learning rate is $\texttt{2e-5}$. The learning rate scheduler has 10\% steps linear warmup then linear decay till 0. We train for 5 epochs and the batch size is 32. For token level tasks, the task-specific layer takes the representation of the first subword, following previous work \citep{devlin-etal-2019-bert,wu-dredze-2019-beto}. Model selection is done on the corresponding dev set of the training set. We fine-tune each model using a single Quadro RTX 6000 and it takes less than one hour except for XNLI.

During fine-tuning, the maximum sequence length is 128. We use a sliding window of context to include subwords beyond the first 128 for NER and POS tagging. At test time, we use the same maximum sequence length with the exception of parsing, where the first 128 words instead of subwords of a sentence were used. We ignore words with POS tags of \texttt{SYM} and \texttt{PUNCT} during parsing evaluation. For NER, the prediction of \texttt{BIO} was post-processed to make sure a valid span is produced. 

All datasets we used are publicly available: NER\footnote{\url{https://www.amazon.com/clouddrive/share/d3KGCRCIYwhKJF0H3eWA26hjg2ZCRhjpEQtDL70FSBN}}, XNLI\footnote{\url{https://dl.fbaipublicfiles.com/XNLI/XNLI-MT-1.0.zip}}\footnote{\url{https://dl.fbaipublicfiles.com/XNLI/XNLI-1.0.zip}}, POS tagging and dependency parsing\footnote{\url{https://lindat.mff.cuni.cz/repository/xmlui/handle/11234/1-3424}}.
For POS tagging and dependency parsing, we use the following treebanks: Arabic-PADT, German-GSD, English-EWT, Spanish-GSD, French-GSD, Hindi-HDTB, Russian-GSD, Vietnamese-VTB, and Chinese-GSD.
Data statistic can be found in \cref{tab:stat}.

\input{result/dataset-stats}

\section{Norm Ratio and Angle of $\delta_{src}$ and $\delta_{tgt}$} \label{sec:dist-vs-angle}

\begin{figure*}[t]
\centering
\includegraphics[width=2\columnwidth]{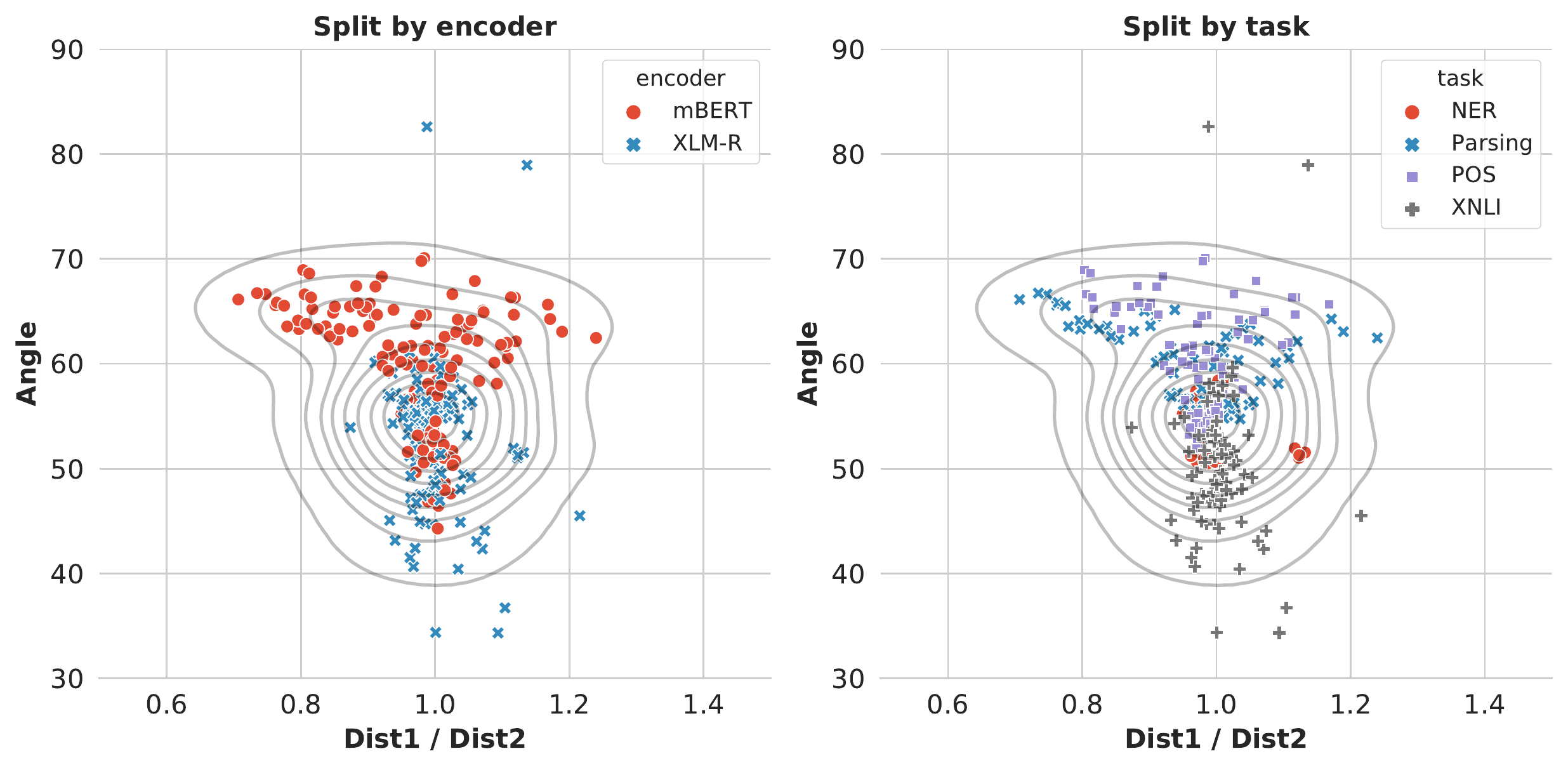}
\caption{
$\|\delta_{src}\|/\|\delta_{tgt}\|$ v.s. angle between $\delta_{src}$ and $\delta_{tgt}$.
Most $\delta_{src}$ and $\delta_{tgt}$ have similar norms, and the angle between them is around 55\degree.
}
\label{fig:dist-vs-angle}
\end{figure*}

\cref{fig:dist-vs-angle} plots the relationship between $\|\delta_{src}\|/\|\delta_{tgt}\|$ and angle between $\delta_{src}$ and $\delta_{tgt}$. We observe most $\delta_{src}$ and $\delta_{tgt}$ have similar norms, and the angle between them is around 55\degree.

\section{Normalized Variance of Linear Interpolated Models}\label{sec:variance}

\begin{figure*}[t]
\centering
\includegraphics[width=2\columnwidth]{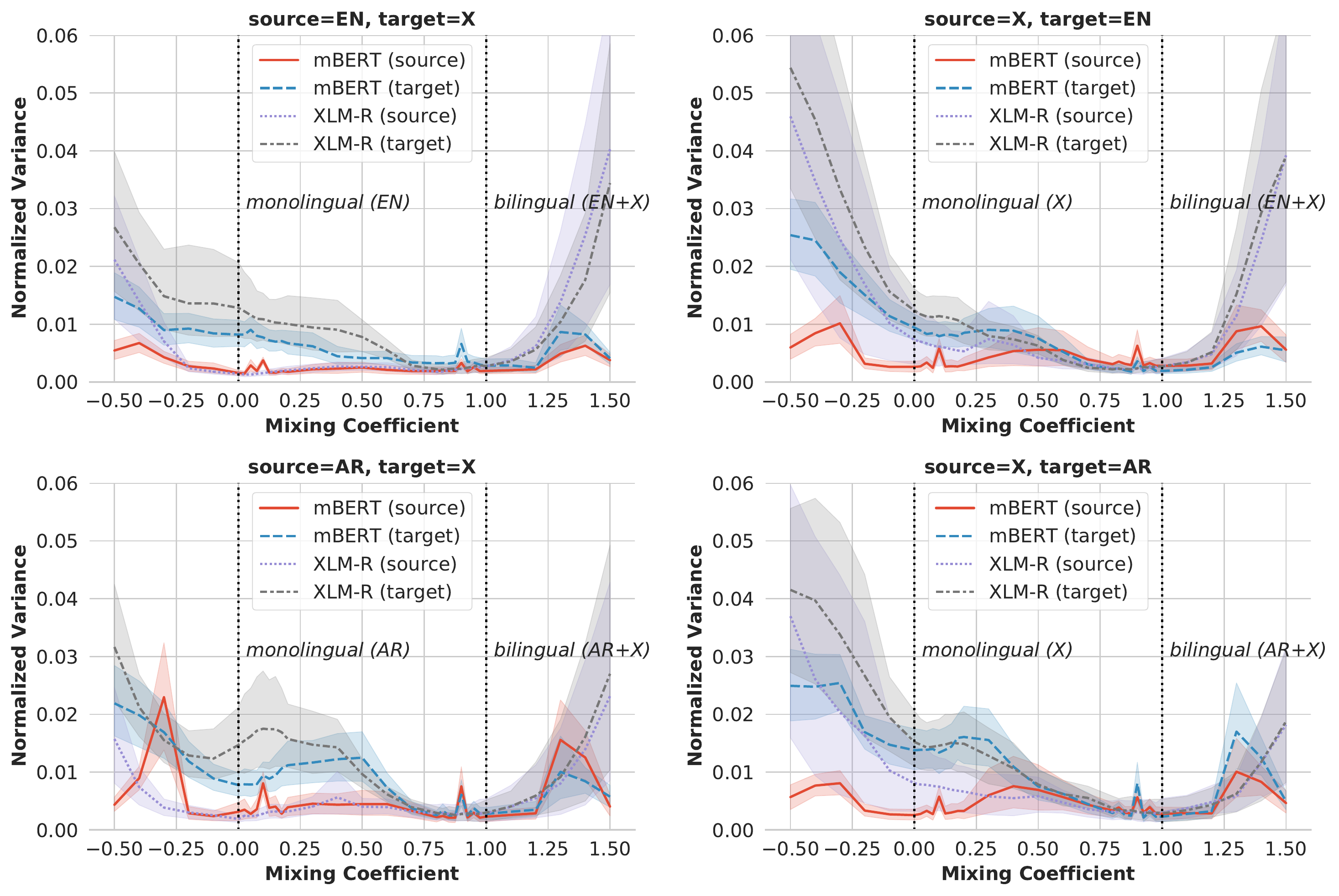}
\caption{
Normalized variance of linear interpolation between monolingual model and bilingual model.
The source language has much lower variance compared to target language on the monolingual side of the interpolated models.}
\label{fig:interpolation-std}
\end{figure*}

\cref{fig:interpolation-std} plots the normalized variance of linear interpolated models.

\section{Break Down of Normalized Performance of Linear Interpolated Models by Tasks}\label{sec:mean-breakdown}

\begin{figure*}[t]
\centering
\includegraphics[width=2\columnwidth]{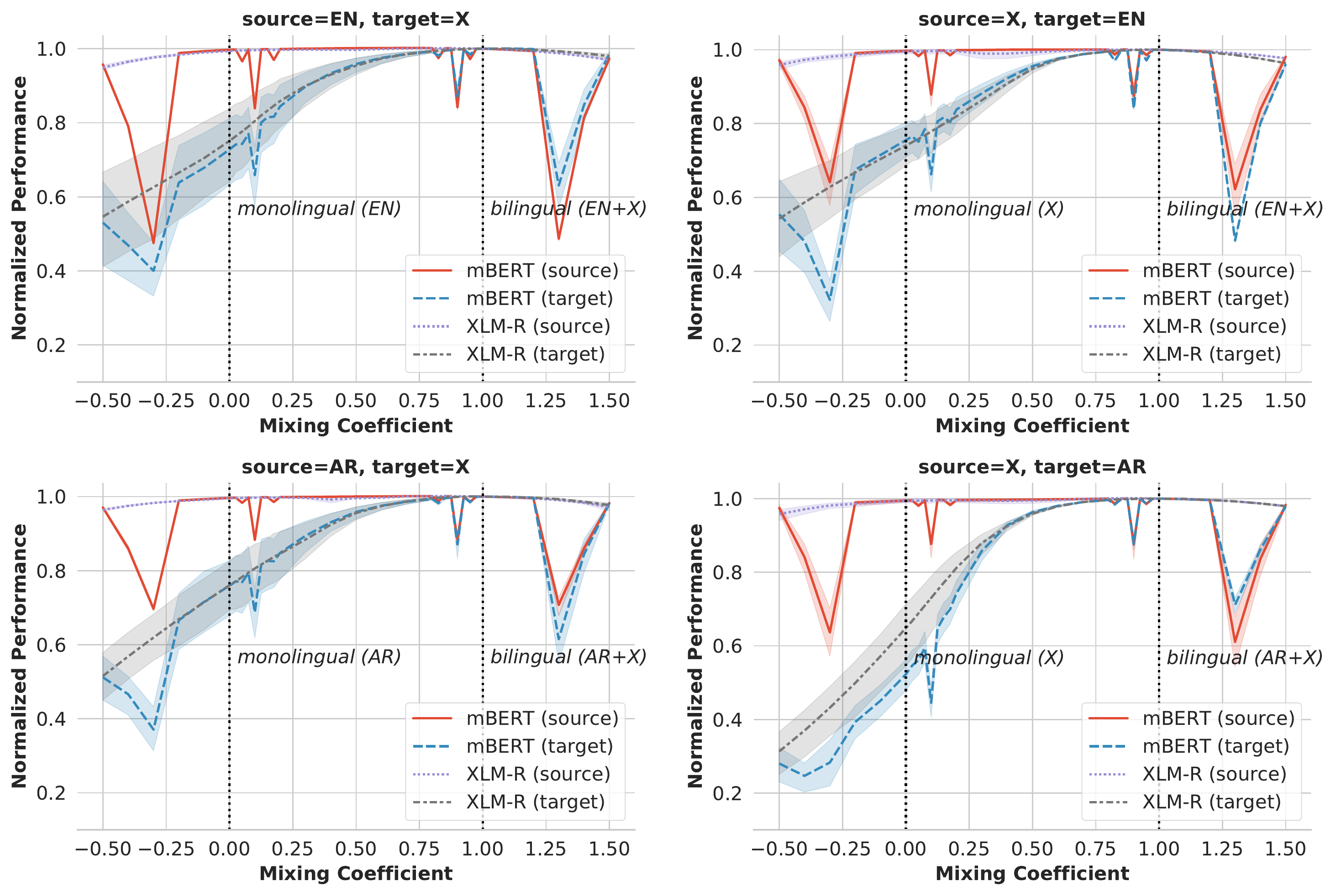}
\caption{Normalized NER performance of linear interpolated model between monolingual and bilingual model}
\label{fig:interpolation-mean-ner}
\end{figure*}

\begin{figure*}[t]
\centering
\includegraphics[width=2\columnwidth]{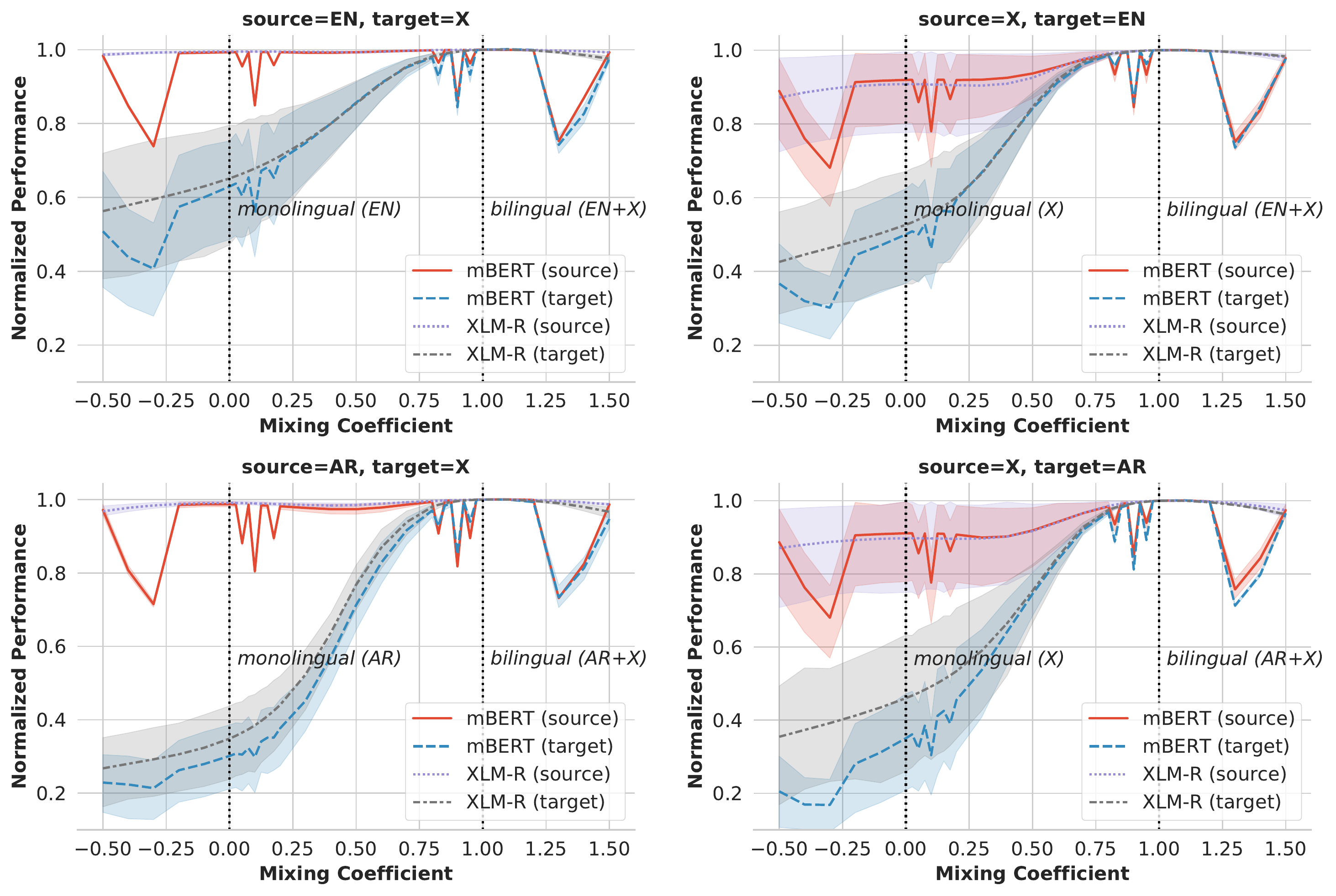}
\caption{Normalized Parsing performance of linear interpolated model between monolingual and bilingual model}
\label{fig:interpolation-mean-parsing}
\end{figure*}

\begin{figure*}[t]
\centering
\includegraphics[width=2\columnwidth]{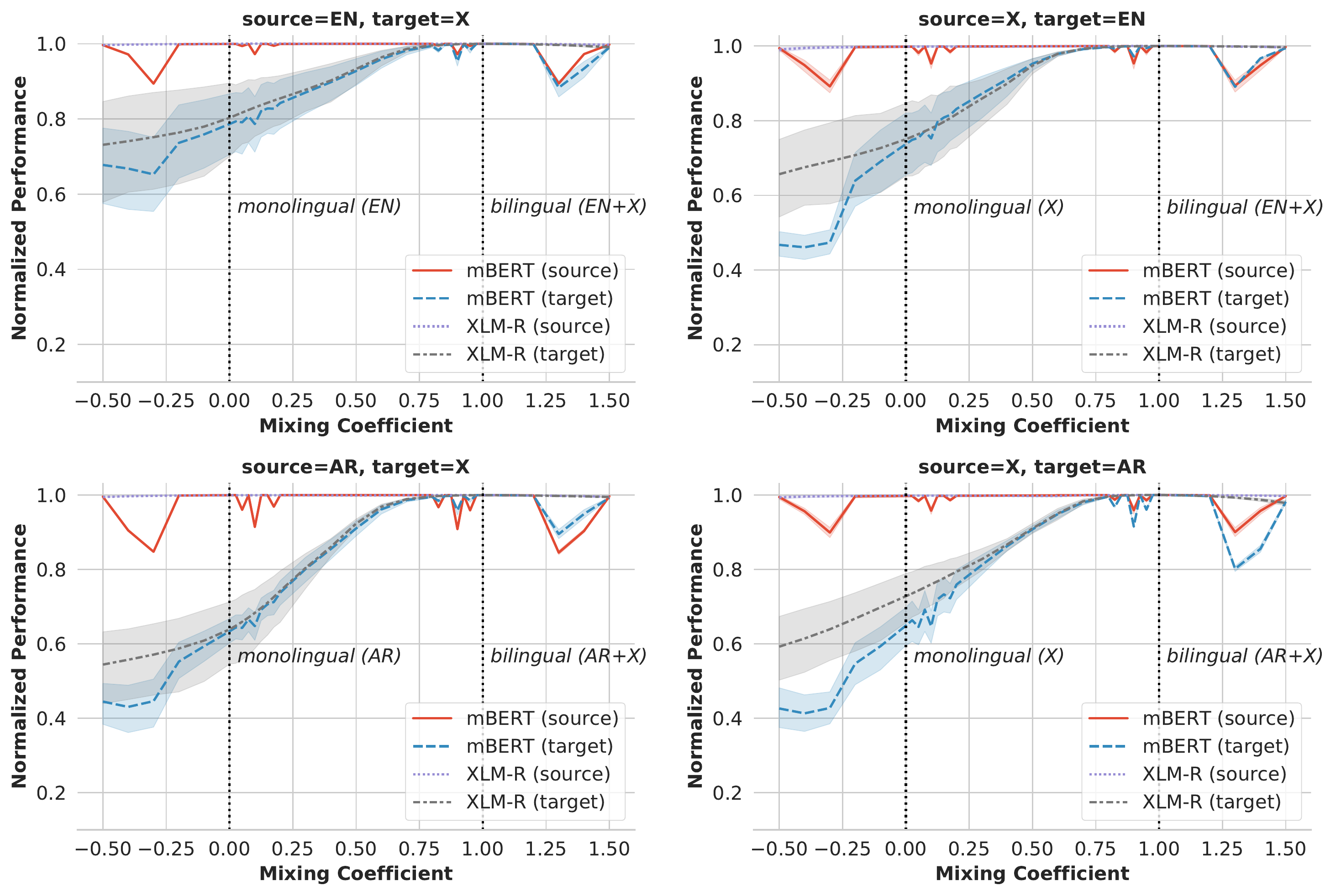}
\caption{Normalized POS performance of linear interpolated model between monolingual and bilingual model}
\label{fig:interpolation-mean-pos}
\end{figure*}

\begin{figure*}[t]
\centering
\includegraphics[width=2\columnwidth]{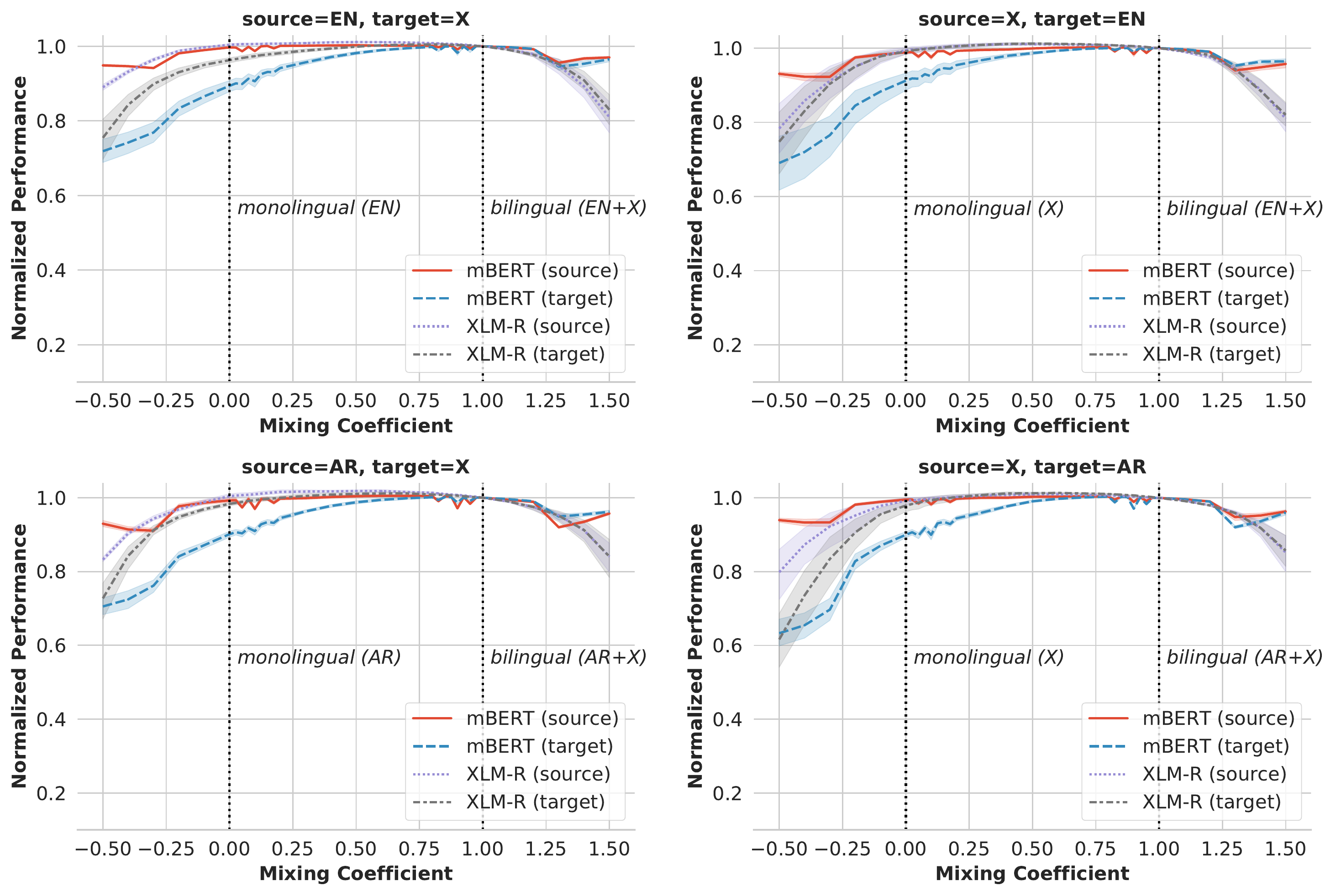}
\caption{Normalized XNLI performance of linear interpolated model between monolingual and bilingual model}
\label{fig:interpolation-mean-xnli}
\end{figure*}

\cref{fig:interpolation-mean-ner} (NER), \cref{fig:interpolation-mean-parsing} (Parsing), \cref{fig:interpolation-mean-pos} (POS), and \cref{fig:interpolation-mean-xnli} (XNLI) plot the normalized performance of linear interpolated models break down by task. We observe similar findings as \cref{fig:interpolation-mean}.

\section{Additional 2D Linear Interpolation}\label{sec:interpolation-2d-others}

\begin{figure*}[t]
\centering
\subfloat[][EN-HI Parsing w/ mBERT]{
\includegraphics[width=0.5\columnwidth]{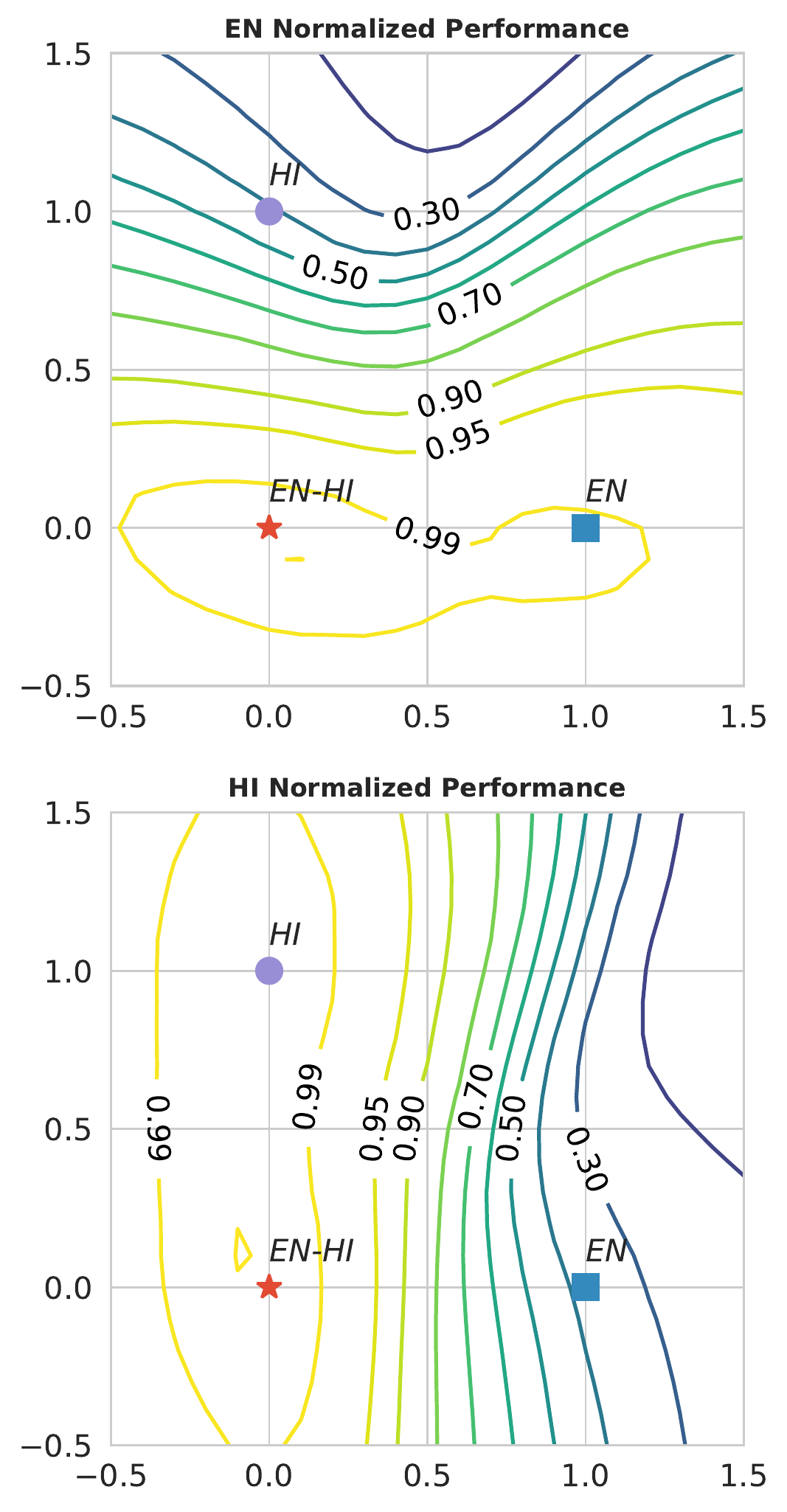}
}
\subfloat[][EN-HI Parsing w/ XLM-R]{
\includegraphics[width=0.5\columnwidth]{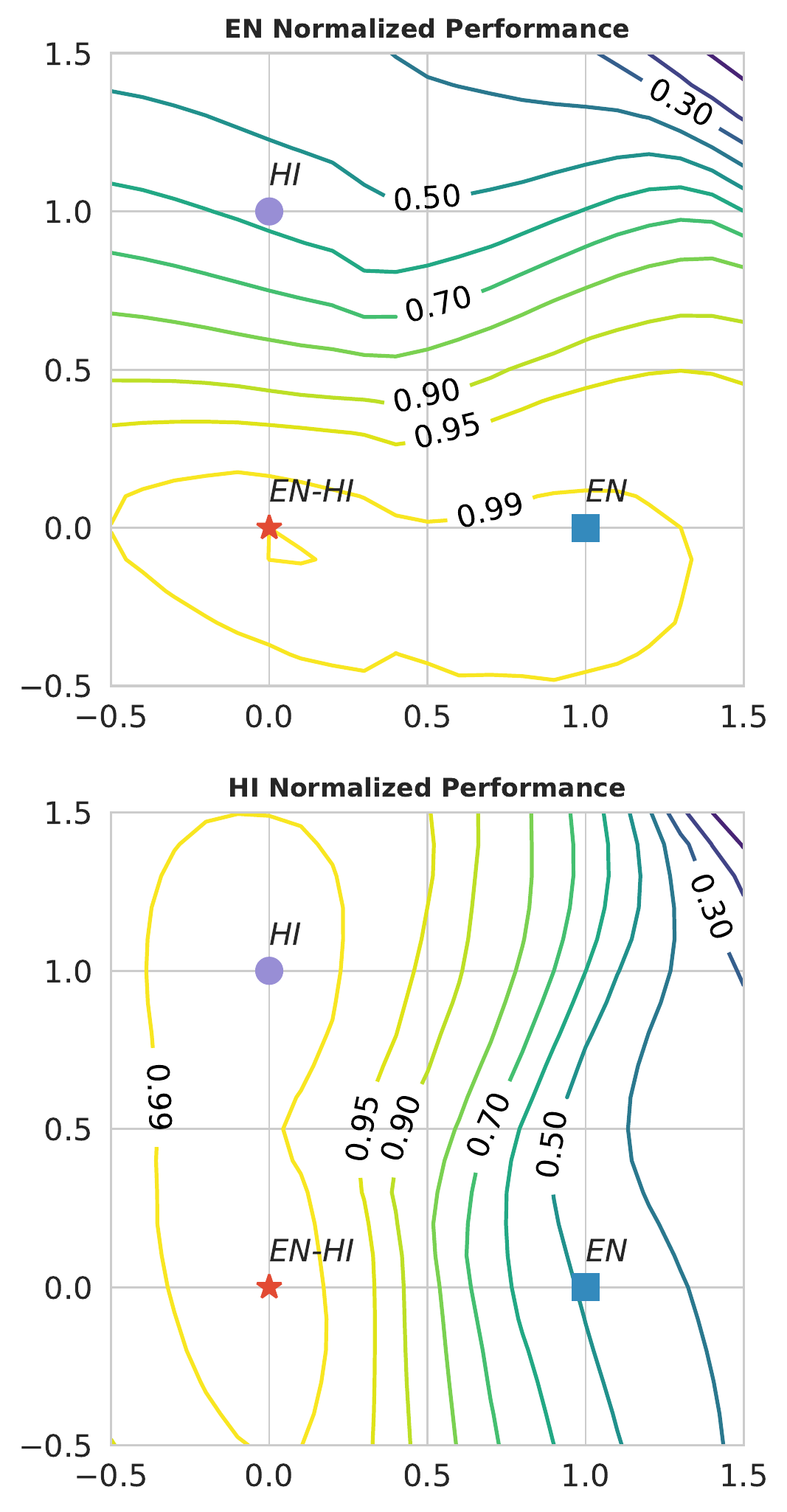}
}
\subfloat[][AR-DE POS w/ mBERT]{
\includegraphics[width=0.5\columnwidth]{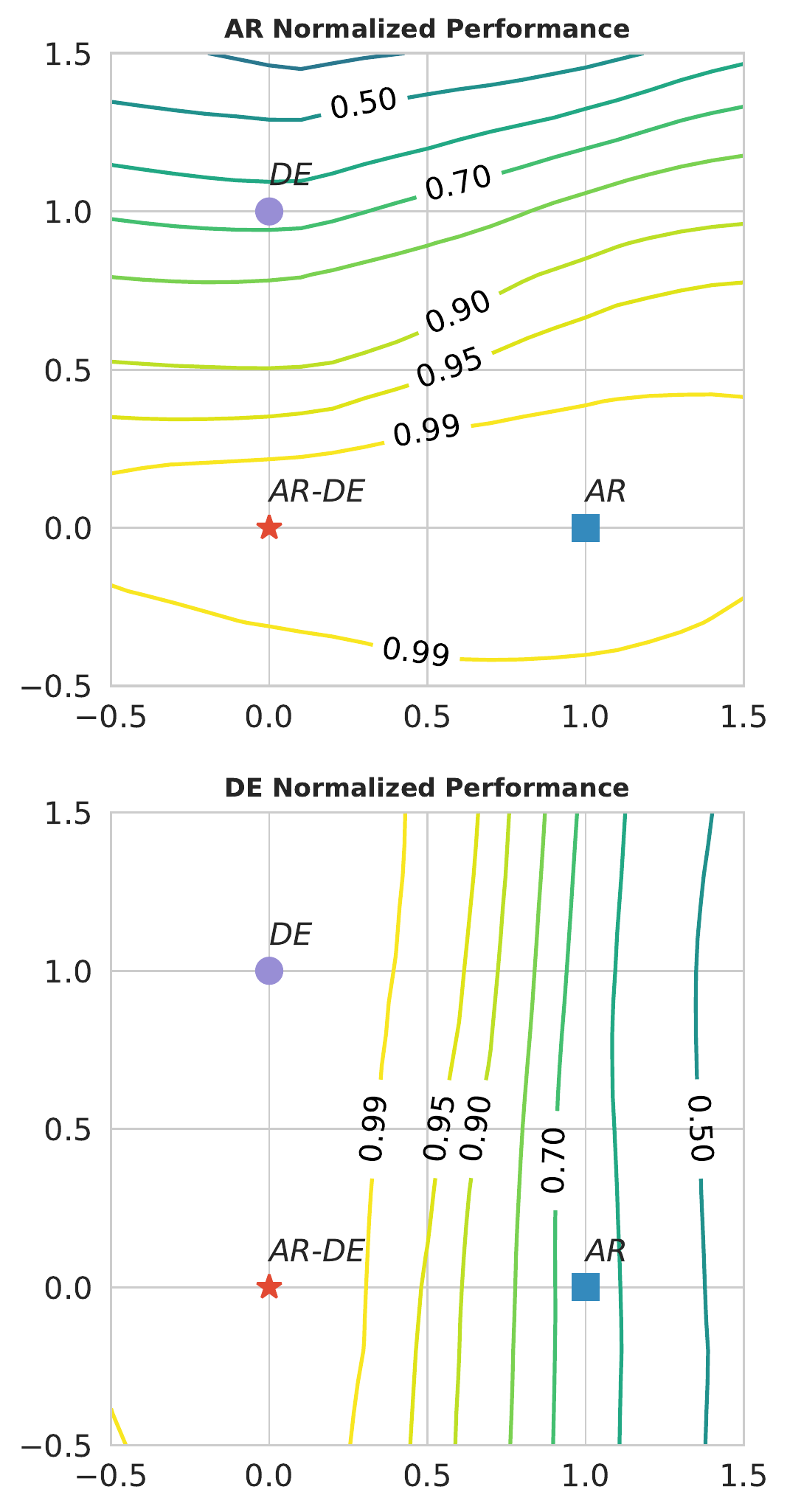}
}
\subfloat[][AR-DE POS w/ XLM-R]{
\includegraphics[width=0.5\columnwidth]{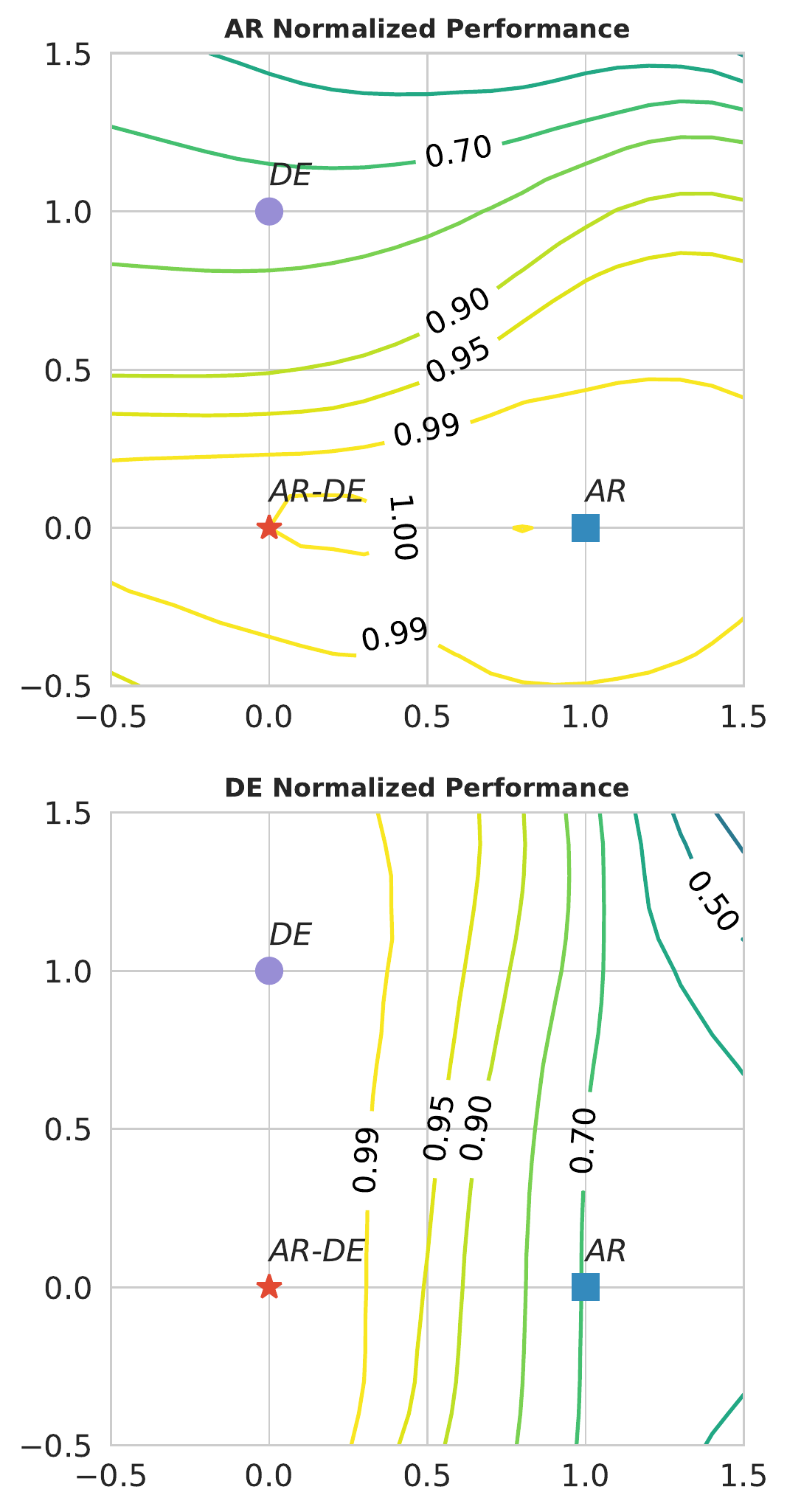}
}

\caption{Additional normalized performance of 2D linear interpolation between bilingual model and monolingual models}
\label{fig:interpolation-2d-others}
\end{figure*}

\cref{fig:interpolation-2d-others} plots additional 2D linear interpolation. We observe similar findings as \cref{fig:interpolation-2d}.

\end{document}

%% file: result/dataset-stats.tex
\begin{table}[h]
\begin{center}
\resizebox{0.8\linewidth}{!}{
\begin{tabular}[b]{lccc}
\toprule
 & \multirow{2}{*}{XNLI} & \multirow{2}{*}{NER} & POS tagging \\
 & & & Parsing\\
\midrule
en-train & 392703 & 20000 & 12543 \\
en-dev & 2490 & 10000 & 2002 \\
\midrule
ar-train & 392703 & 20000 & 6075 \\
ar-dev & 2490 & 10000 & 909 \\
\midrule
de-train & 392703 & 20000 & 13814 \\
de-dev & 2490 & 10000 & 799 \\
\midrule
es-train & 392703 & 20000 & 14187 \\
es-dev & 2490 & 10000 & 1400 \\
\midrule
fr-train & 392703 & 20000 & 14449 \\
fr-dev & 2490 & 10000 & 1476 \\
\midrule
hi-train & 392703 & 5000 & 13304 \\
hi-dev & 2490 & 1000 & 1659 \\
\midrule
ru-train & 392703 & 20000 & 3850 \\
ru-dev & 2490 & 10000 & 579 \\
\midrule
vi-train & 392703 & 20000 & 1400 \\
vi-dev & 2490 & 10000 & 800 \\
\midrule
zh-train & 392703 & 20000 & 3997 \\
zh-dev & 2490 & 10000 & 500 \\
\bottomrule
\end{tabular}
}
\caption{Number of examples.\label{tab:stat}}
\end{center}
\end{table}